\documentclass[conference]{IEEEtran}
\IEEEoverridecommandlockouts
\usepackage{cite}
\usepackage{amsmath,amssymb,amsfonts}
\usepackage{algorithmic}
\usepackage{graphicx}
\usepackage{textcomp}
\usepackage{xcolor}
\def\BibTeX{{\rm B\kern-.05em{\sc i\kern-.025em b}\kern-.08em
    T\kern-.1667em\lower.7ex\hbox{E}\kern-.125emX}}

\usepackage[hidelinks]{hyperref}
\usepackage{pgfplots}
\pgfplotsset{width=8cm,compat=1.9}
\usepackage{babel}

\begin{document}

\title{Unsupervised extraction, labelling and clustering of segments from clinical notes\\
{}
\thanks{This work has been supported by the Grant Agency of Masaryk University under grant no. MUNI/G/1763/2020 (AIcope). \newline

\vspace{-5pt} \noindent
978-1-6654-6819-0/22/\$31.00 ©2022 IEEE  Personal use of this material is permitted.  Permission from IEEE must be obtained for all other uses, in any current or future media, including reprinting/republishing this material for advertising or promotional purposes, creating new collective works, for resale or redistribution to servers or lists, or reuse of any copyrighted component of this work in other works.

}
}

\author{\IEEEauthorblockN{1\textsuperscript{st} Petr Zelina}
\IEEEauthorblockA{\textit{Faculty of Informatics,} \\
\textit{Masaryk University}\\
Brno, Czechia \\
\url{469366@mail.muni.cz}}
\and
\IEEEauthorblockN{2\textsuperscript{nd} Jana Halámková}
\IEEEauthorblockA{
\textit{Masaryk Memorial Cancer Institute} \\
Brno, Czechia \\
\url{jana.halamkova@mou.cz}}
\and
\IEEEauthorblockN{3\textsuperscript{rd} Vít Nováček}
\IEEEauthorblockA{
\textit{$^1$Faculty of Informatics, Masaryk University} \\
\textit{$^2$Masaryk Memorial Cancer Institute} \\
\textit{$^3$Data Science Institute, NUI Galway} \\
$^{1,2}$Brno, Czechia, $^{3}$Galway, Ireland\\
\url{novacek@fi.muni.cz}}
}

\maketitle

\begin{abstract}
This work is motivated by the scarcity of tools for accurate, unsupervised information extraction from unstructured clinical notes in computationally underrepresented languages, such as Czech. We introduce a stepping stone to a broad array of downstream tasks such as summarisation or integration of individual patient records, extraction of structured information for national cancer registry reporting or building of semi-structured semantic patient representations for computing patient embeddings. More specifically, we present a method for unsupervised extraction of semantically-labelled textual segments from clinical notes and test it out on a dataset of Czech breast cancer patients, provided by Masaryk Memorial Cancer Institute (the largest Czech hospital specialising in oncology). Our goal was to extract, classify (i.e. label) and cluster segments of the free-text notes that correspond to specific clinical features (e.g., family background, comorbidities or toxicities). The presented results demonstrate the practical relevance of the proposed approach for building more sophisticated extraction and analytical pipelines deployed on Czech clinical notes.

\end{abstract}


\section{Introduction}
Electronic Health Records are notoriously difficult to analyse by automated means as they contain large amounts of unstructured textual contents. This is particularly pertinent to clinical notes, which are narrative reports of clinicians documenting prescriptions and results of a broad array of diagnostic and therapeutic procedures. The challenge of large-scale automated processing of clinical notes is further exacerbated when they are written in a language for which virtually no tools for biomedical Natural Language Processing are available (such as Czech).

\begin{figure}[]
\centering
\vspace{-15pt}
\includegraphics[width=0.8\linewidth]{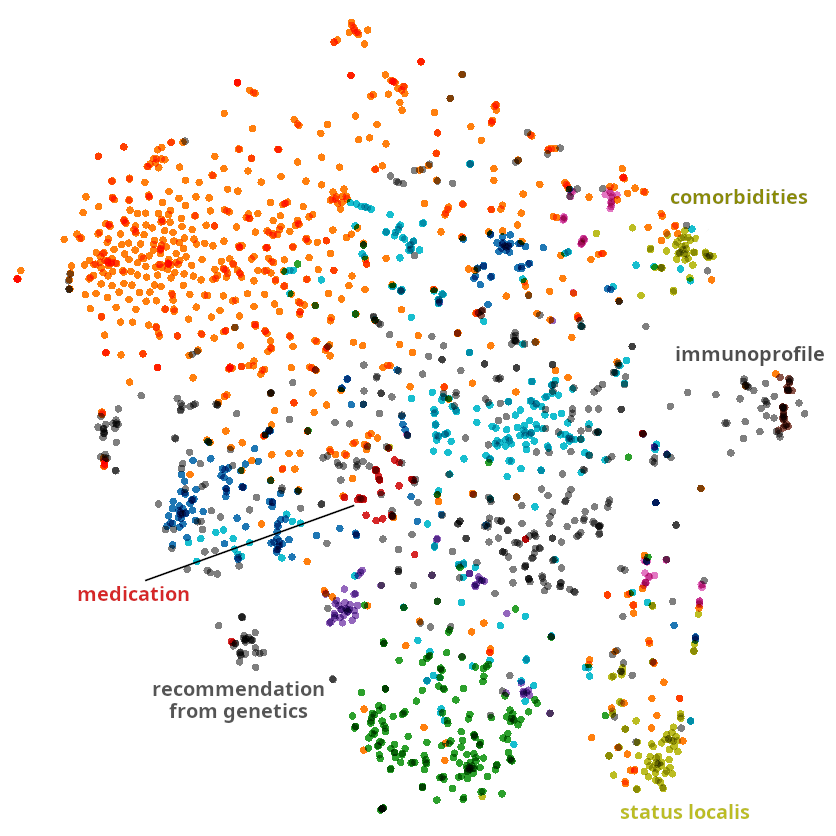}
\setlength{\intextsep}{-20pt}
\caption{Example of title clustering results (K-means $k=20$). Visualised with TensorFlow Embedding Projector and 2D t-SNE. Includes color-coded manually translated labels for some clusters. Interactive version on \href{https://zepzep.github.io/clinical-notes-extraction/pages/projector/}{\textit{GitHub}}.}
\label{fig:title-clustering}
\end{figure}

This paper presents a solution to a specific instance of this problem: we introduce a system for unsupervised extraction of textual segments from clinical notes on Czech breast cancer patients, provided by Masaryk Memorial Cancer Institute (MMCI, or MOU in Czech). Our goal was to extract, classify (i.e. label) and cluster segments of the free-text notes that correspond to specific clinical features (e.g., family background, comorbidities or toxicities). 


As this is the first work we know of in this area, we tested out several methods for classification to develop and compare increasingly sophisticated solutions to the problem. The first two are based on two different segment embedding methods (LSA or Doc2Vec) and subsequent neural classification. The third one uses an end-to-end recurrent neural architecture (Bi-directional LSTM). Finally we experimented with adapting a general-purpose Czech transformer language model (RobeCzech) to this task. The third and fourth models outperform the first two by a rather big margin, reaching an accuracy of over 0.94 which is very promising for downstream real-world clinical applications. Also, note that most of our models are not specifically dependent on the Czech language features and thus they can be adapted to clinical notes in other countries as well from the source code provided on \href{https://github.com/ZepZep/clinical-notes-extraction}{GitHub}\footnote{\href{https://github.com/ZepZep/clinical-notes-extraction}{https://github.com/ZepZep/clinical-notes-extraction}}. The only exception is the classification module based on RobeCzech, a Czech language model, but that specific part of the pipeline can be replaced by one of the many available localised language model variants~\cite{kassner2021multilingual}.

The classified segments were further clustered to generate a more focused representation of the domain, as the meanings of some of the labels were rather close to each other based on clinical expert feedback. We performed a preliminary evaluation of the acquired clusters and designed a system for mapping the labels to existing healthcare ontologies. For an example of the final result of one specific configuration of the clustering module, see Figure \ref{fig:title-clustering}.

Our results demonstrate the practical potential of the presented approach for downstream analytical and information extraction tasks, such as integration or summarisation of individual patient records, extraction of structured information for national cancer registry reporting or building of semi-structured semantic patient representations for computing patient embeddings.




\section{Related Work}

Works related to the presented research can be broadly divided into supervised and unsupervised approaches. In terms of the supervised ones, \cite{li2010section} classifies sections of clinical notes into 15 predefined categories using hidden Markov models. The work~\cite{ganesan2014general} focuses on clinical notes following the SOAP (Subjective, Objective, Assessment and Plan) structure and can learn the segment labels directly from the text via logistic regression model. The approach in~\cite{edinger2017evaluation} uses segmentation based on hand-crafted rules. A machine learning approach based on bag-of-words feature modelling, classifying segments to 5 pre-defined categories, is introduced in~\cite{ruan2018boundary}.

In terms of the unsupervised approaches, \cite{wen2021mining} uses latent topic models to jointly infer distinct topic distributions of notes of different types. The work~\cite{alicante2014unsupervised} did unsupervised segmentation of 57 Italian records, focused rather on entity and relation extraction using domain knowledge (dictionaries) and language features (part of speech).

The supervised approaches typically focus on a limited set of predefined categories into which the textual segments can be classified, and/or rely on labour-intensive feature engineering. These are limitations our approach addresses better than the referenced works.

Among the unsupervised methods, the works presented in~\cite{wen2021mining,fette2012information} are conceptually closer to our approach than the third referenced work focused on entity and relation extraction from the segments. However, \cite{wen2021mining} is specifically tuned towards ICU clinical notes, which are rather different from the more general breast cancer clinical notes we work with, and \cite{fette2012information} is based on manually designed taxonomies for each type of note and thus not entirely unsupervised.

It should be noted that some papers, such as~\cite{ganesan2014general} or~\cite{ruan2018boundary}, use the $P_k$ metric for measuring the segmentation performance. While being a de facto standard in this task, we do not use it in this work as our segmentation algorithm is unsupervised and thus we do not posses ground truth segmentation required to compute the $P_k$ metric.

Two more papers of note are the following. The work~\cite{weng2017medical} is similar to ours in terms of using deep (representation) learning based on distributional semantics, but it focuses rather on medical subdomains as labels of text segments, not bottom-up label generation from the segments themselves as in our case. Finally, the work~\cite{patrick2011knowledge} is an example of a full-fledged information extraction pipeline deployed on clinical notes in clearly defined and standardised benchmark settings (the i2b2/VA challenge), which is an approach we intend to further explore for downstream applications of our system.

\section{Methods}
Our approach consists of several steps, which can be customised in order to accommodate the specifics in each dataset.

In order to obtain the classifier and clusters we first need to create the training data. We split the medical records into segments and extract the segment labels. Then we can use this dataset to create a classifier for segment types or apply some clustering methods to group similar labels or map them to existing ontologies. 

\begin{figure}[h]
\centering
\includegraphics[width=\linewidth,]{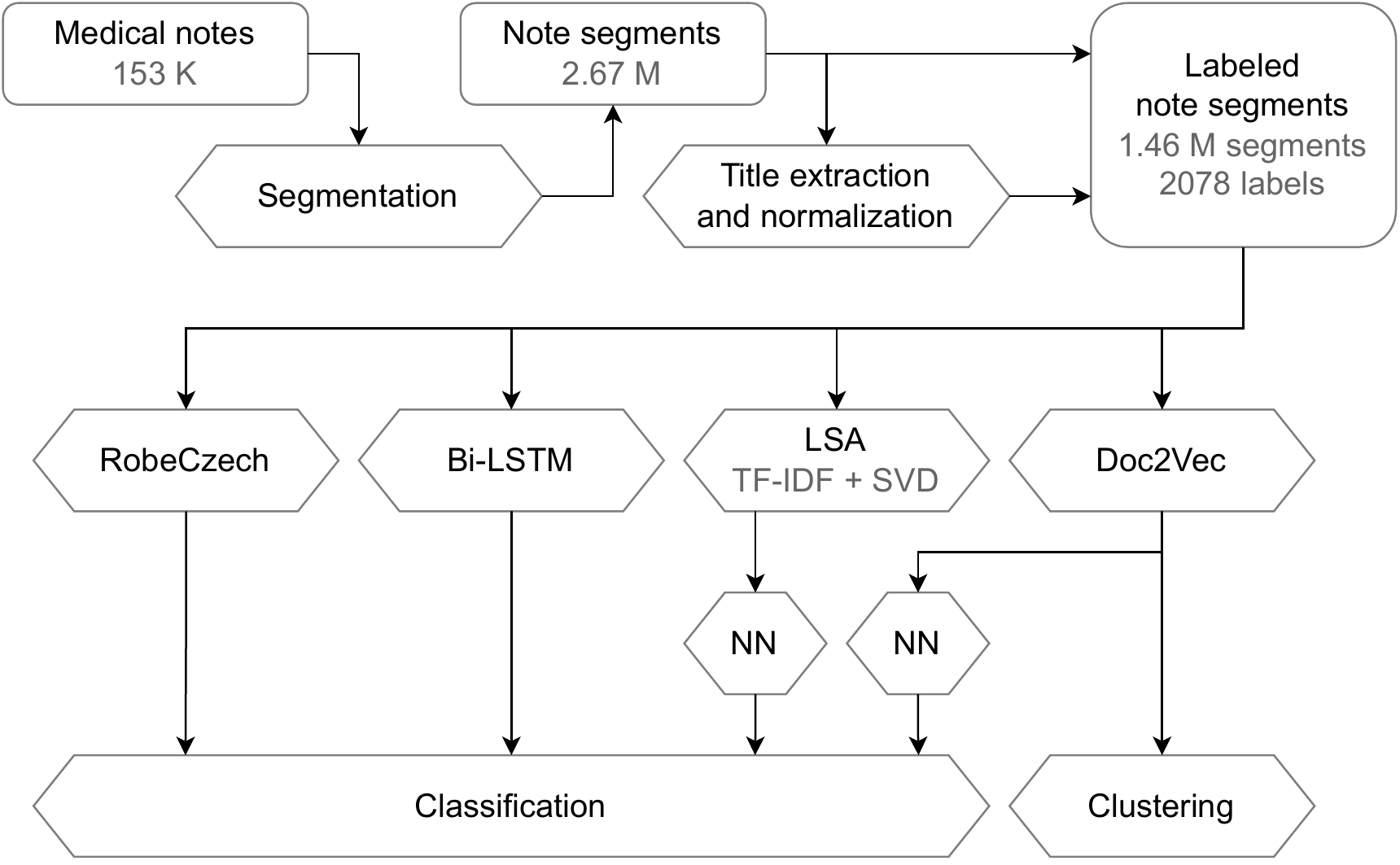}
\caption{Overview of our approach. \textit{Rounded rectangles:} data and their size. \textit{Hexagons:} procedures. We explored 4 classification techniques.}
\label{fig:overview}
\end{figure}

Figure \ref{fig:overview} shows an overview of our approach.  More details for each step are provided below. 

\subsection{Record Splitting}
In order to split the medical records into segments we used a hard-coded splitting algorithm. This step may require some fine-tuning specifically for the given language, hospital or field of medicine, because the formatting may differ significantly. 

We noticed some repeating patterns characteristic to our MMCI breast cancer patient data and used them for the splitting:
\begin{itemize}
    \item segments often start with titles (labels) in the format of \textit{title: \dots}
    \item new line ends a segment, unless the previous line contains only a title, or the next line is either indented or contains a dash or a bullet point.
    \item empty lines always denote an end of a segment
\end{itemize}

The exact algorithm can be seen in the GitHub repository\footnote{\href{https://github.com/ZepZep/clinical-notes-extraction}{https://github.com/ZepZep/clinical-notes-extraction} \newline function \texttt{cut\_record} inside the \texttt{make\_segments.py} file}.

Some lines contain multiple short segments (e.g. weight, height, blood pressure), but the delimiters were too inconsistent and ambiguous (semicolons, commas or even just spaces) for reliable segmentation. Furthermore because the segments are short (e.g. single number) there is too little context for the models to predict the title. For these reasons we did not attempt to split such parts of the texts.

\subsection{Label Extraction}
Many segments contain titles (labels) like Date, Diagnosis, Case history, Recommendation, etc. We are able to extract these titles with a simple regular expression because they consistently start at the beginning and end with a colon. This way we obtain a labeled dataset of medical record segments.

Because there is too many unique labels, we further normalize the label text (convert to lowercase, remove accents and trailing spaces). We also discard the normalized labels which contain too many words (more than 4) or which occur less than 10 times in the dataset, as these were deemed to be insignificant outliers based on our discussions with the clinical experts.

\subsection{Segment Classification}
The goal of this part of the system is to predict the segment title (label) based only on the contents of the segment. To prevent implicit leak of labels into the classified text chunks, we want to classify two possible representations of the segments: with or without the title. Thus we enhance the dataset by adding all segments again, but with their titles removed from the text. We make sure to split the dataset into train / test before this procedure in order to avoid explicit data leaks.

Now we can use any model for sequence classification. We tried out the following models:
\begin{itemize}
    \item most frequent label dummy baseline
    \item LSA + simple neural network classifier
    \item Doc2Vec + simple neural network classifier
    \item Bi-LSTM neural network classifier
    \item A transformer based on the RoBERTa architecture, pre-trained on Czech texts
\end{itemize}

More details about the specific configurations of these models are provided in the Experiments section.

\subsection{Title Clustering}


Our goal was to extract a small number (e.g. 50) of meta-titles which would nicely and uniformly (not everything in a single meta-title) cover all the titles/segments. But it is not clear how to evaluate the usefulness of such clustering without a downstream task and a corresponding gold standard that does not exist yet for our specific use case.

The first step is to create embeddings of the titles. We chose to use the Doc2Vec embeddings, because they directly output the document (in our case title) embedding. Other methods provide embeddings for individual segments and we would need to somehow average over all segments with a given title.

Once we have the title embeddings we can apply usual clustering algorithms.

Because there is no clear metric for evaluating the resulting clusters we focused more on the vector space created by the embedding. We are able to look at similar titles or measure distances between titles.

We can also use the title distance matrix and title frequency for computer assisted ontology fitting. If we want to sort the note segments into a given ontology (e.g. LOINC Consult note\footnote{\href{https://loinc.org/11488-4.html/}{https://loinc.org/11488-4.html/}}), we just need to make a mapping from our extracted normalized titles to the ontology entries.
We can create a simple recommendation system that suggests similar frequent titles based on the title distance matrix and title frequencies.
Given the long tail nature of title frequencies (see section Experiments and Figure \ref{fig:label-distribution}) it should be enough to manually assign a small amount of titles (hundreds) for high coverage. Figure \ref{fig:title-sims} in the Experiments section demonstrate the feasibility of this approach, which we want to further explore and validate in our future work with the MMCI clinicians.

\subsection{Evaluation}
\label{sec:evaluation}

\subsubsection{Segment classification}
We evaluate the classifiers with a stratified test set consisting of 20 \% of the data.
We look at 5 metrics:
\begin{itemize}
    \item \textit{accuracy} (acc) -- fraction of correctly predicted labels
    \item \textit{macro F1} (MF1) -- arithmetic mean of F1 scores
    \item \textit{weighted F1} (wF1) -- mean of F1 scores weighted by the class frequency (also known as micro F1)
    \item \textit{top-5 accuracy} (acc@5) -- fraction of predictions which contained the correct label in the top 5 predicted labels
    \item \textit{top-10 accuracy} (acc@10) -- fraction of predictions which contained the correct label in the top 10 predicted labels
\end{itemize}

We also perform a small manual review of the predictions for segments from which we were not able to extract the title and frequent titles which have low F1 score. 

\subsubsection{Title clustering}
Evaluation of clustering without a downstream task is difficult. In the future we plan on using this system for measuring patient similarity but this is still in progress. Therefore we only include some illustrative examples and perform only a preliminary qualitative assessment of this task here.

\section{Experiments}

\subsection{Dataset}
For our experiments we used a Czech dataset of breast cancer patients provided by the Masaryk Memorial Cancer Institute\footnote{\href{https://www.mou.cz/en/}{https://www.mou.cz/en/}, Brno, Czechia}. Due to privacy, ethical and legal concerns, we cannot share this dataset, but can provide more details to interested parties on request.

In total we obtained medical records from 4267 patients ($\sim$300 MB of raw text). The dataset contains 153~K records, yielding $\sim$36 records per patient on average.

\subsection{Extraction Pipeline}
Our splitting algorithm results in 2.67~M segments ($\sim$18~segments per record). We were able to extract titles from 1.58~M segments (59\%). 

There were 60~K unique titles which we cut down to only 2078 by normalization and thresholding (at least 10 occurrences after normalization). The final 2078 titles cover 1.46~M segments (92\% of the labeled segments).

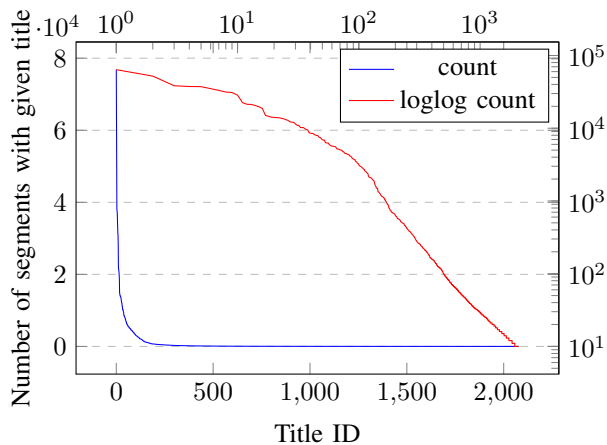
\begin{figure}[]
\centering
\begin{tikzpicture}
\begin{axis}[
    xlabel={Title ID},
    ylabel={Number of segments with given title},
    height=6cm,
    ymajorgrids=true,
    grid style=dashed,
    every y tick scale label/.style={at={(yticklabel cs:1.05)}, xshift=0.5em},
]

\addplot table [x=index, y=count, col sep=comma, mark=none] {data/label-dist.csv};
\label{plot_one}
\end{axis}

\begin{axis}[
    axis y line*=right,
    axis x line*=top,
    height=6cm,
    ymode=log,
    xmode=log,
    xtick={1,10, 100, 1000},
    cycle list name = color list,
]
\addlegendimage{blue}\addlegendentry{count}
\addplot table [red, x=index, y=count, col sep=comma, mark=none] {data/label-dist.csv};
\addlegendentry{loglog count}
\end{axis}
\end{tikzpicture}

\caption{Distribution of the count of segments with given normalized titles, sorted from the most common. For clarity we include both normal (bottom left and blue) and loglog (top right and red) scale.}
\label{fig:label-distribution}
\end{figure}

Figure \ref{fig:label-distribution} shows the distribution of the title counts. As we can see it has a long tail, meaning there is a lot of titles which appear only in a small number of segments (Zipf's law).

\subsection{Classification Pipelines}

\subsubsection{Most frequent label dummy baseline (Top label)} 
This model is a simple baseline which always guesses the most common class (or all classes sorted by their frequency for the top-n accuracy).

\subsubsection{LSA + simple neural network classifier (LSA)} This model consists of two parts: First, Latent Semantic Analysis (LSA) \cite{lsa} (TF-IDF segment vectorization followed by SVD dimensionality reduction) which results in segment embeddings. Second, we train a simple dense neural network to predict the label of the segment based on the segment embedding.

We chose 50-dimensional embedding (there are only 2078 document types). The neural network has 1 hidden layer with 64 neurons and ReLU activation. We used the Adam optimizer and trained for 10 epochs. 

\subsubsection{Doc2Vec + simple neural network classifier (Doc2Vec)} Similarly as the previous method, this model consists of two parts: First we train Doc2Vec embeddings \cite{doc2vec} on the train segments, using the labels as document IDs. Doc2Vec primarily outputs the document (i.e. title) embeddings but it is also able to vectorize segments. We use this feature to infer all the segment embeddings. Second we train a simple dense neural network to predict the label of the segment based on the segment embedding.

As in the previous method we use 50-dimensional embedding and the same hyperparameters for the neural network.

\subsubsection{Bi-LSTM neural network classifier (Bi-LSTM)} This model is a single neural network that consists of an embedding layer, bidirectional LSTM layer \cite{bilstm} and a dense classification layer.

Because this is a recurrent neural network we needed to pad~/~cut all segments to the same number of tokens. Based on the distribution of the train set we chose to pad to 150 tokens. We used the same tokenizer that the RobeCzech model uses (see next model). The token embedding dimension is 128 and the Bi-LSTM layer has 2x300 units. In total the network has approx. 9 M parameters. We used the Adam optimizer and trained for 10 epochs.

\subsubsection{Fine-tuned RoBERTa based transformer (RobeCzech)} The final model uses a big pre-trained Czech transformer called RobeCzech \cite{robeczech}. It is based on the RoBERTa  \cite{roberta} architecture. We added a sequence classification head on top of the pre-trained model (connected to the \texttt{[CLS]} token) and fine-tuned it on our training dataset.

Because RobeCzech is a pre-trained model, most of the hyperparameters are not configurable. The segments are padded~/~cut to 512 tokens. We use a cumulative batch size of 128, mixed precision and fine-tune for 10 epochs without freezing the lower layers. Because this model is quite big (125~M parameters), the fine-tuning took around 3 days.

\subsection{Classification Results}

Our results are summarized in Table \ref{tab:classification-results}. We chose to measure \textit{acc} and \textit{wF1} to see how the classifiers perform on an ``average segment'', \textit{MF1} to see how the classifiers perform on an ``average class'' (title) and \textit{acc@5} and \textit{acc@10} to see what is the benefit of looking at more than just the top prediction. More details about the chosen metrics are in the Methods/Evaluation Section (\ref{sec:evaluation}). 

\begin{table}[htbp]
\caption{Classification results}
\begin{center}
\begin{tabular}{|r|c|c|c|c|c|}

\hline
method & acc & MF1 & wF1 & acc@5 & acc@10 \\
\hline

Top label & 0.05 & 0.00 & 0.01 & 0.18 & 0.30 \\
LSA & 0.73 & 0.09 & 0.71 & 0.87 & 0.90 \\
Doc2Vec & 0.38 & 0.05 & 0.40 & 0.48 & 0.50 \\
Bi-LSTM & 0.94 & 0.82 & 0.94 & 0.98 & \textbf{0.99} \\
RobeCzech & \textbf{0.95} & \textbf{0.86} & \textbf{0.95} & \textbf{0.99} & \textbf{0.99} \\
\hline

\end{tabular}
\label{tab:classification-results}
\end{center}
\end{table}

Because we wanted the models to be able to classify both segments with and without a title, we used the segments twice.
One half of the dataset consists of the original segments (with title) and in the second half we removed the titles. We made sure to split the dataset into train~/~test before this procedure to avoid data leaking. Tables \ref{tab:classification-results-titles} and \ref{tab:classification-results-notitles} show evaluation on both parts of the test set separately.

\begin{table}[htbp]
\caption{Classification results -- only segments with titles \textbf{included}}
\begin{center}
\begin{tabular}{|r|c|c|c|c|c|}

\hline
method & acc & MF1 & wF1 & acc@5 & acc@10 \\
\hline

Top label & 0.05 & 0.00 & 0.01 & 0.18 & 0.30 \\
LSA & 0.81 & 0.10 & 0.78 & 0.91 & 0.93 \\
Doc2Vec & 0.43 & 0.06 & 0.45 & 0.53 & 0.55 \\
Bi-LSTM & \textbf{1.00} & 0.97 & \textbf{1.00} & \textbf{1.00} & \textbf{1.00} \\
RobeCzech & \textbf{1.00} & \textbf{0.99} & \textbf{1.00} & \textbf{1.00} & \textbf{1.00} \\
\hline

\end{tabular}
\label{tab:classification-results-titles}
\end{center}
\end{table}

\begin{table}[htbp]
\caption{Classification results -- only segments with titles \textbf{removed}}
\begin{center}
\begin{tabular}{|r|c|c|c|c|c|}

\hline
method & acc & MF1 & wF1 & acc@5 & acc@10 \\
\hline

Top label & 0.05 & 0.00 & 0.01 & 0.18 & 0.30 \\
Doc2Vec & 0.32 & 0.04 & 0.33 & 0.42 & 0.46 \\
LSA & 0.65 & 0.07 & 0.62 & 0.83 & 0.87 \\
Bi-LSTM & 0.88 & 0.61 & 0.87 & 0.97 & 0.98 \\
RobeCzech & \textbf{0.90} & \textbf{0.68} & \textbf{0.90} & \textbf{0.98} & \textbf{0.99} \\
\hline

\end{tabular}
\label{tab:classification-results-notitles}
\end{center}
\end{table}

\begin{figure*}[]
\centering
\includegraphics[width=\linewidth,]{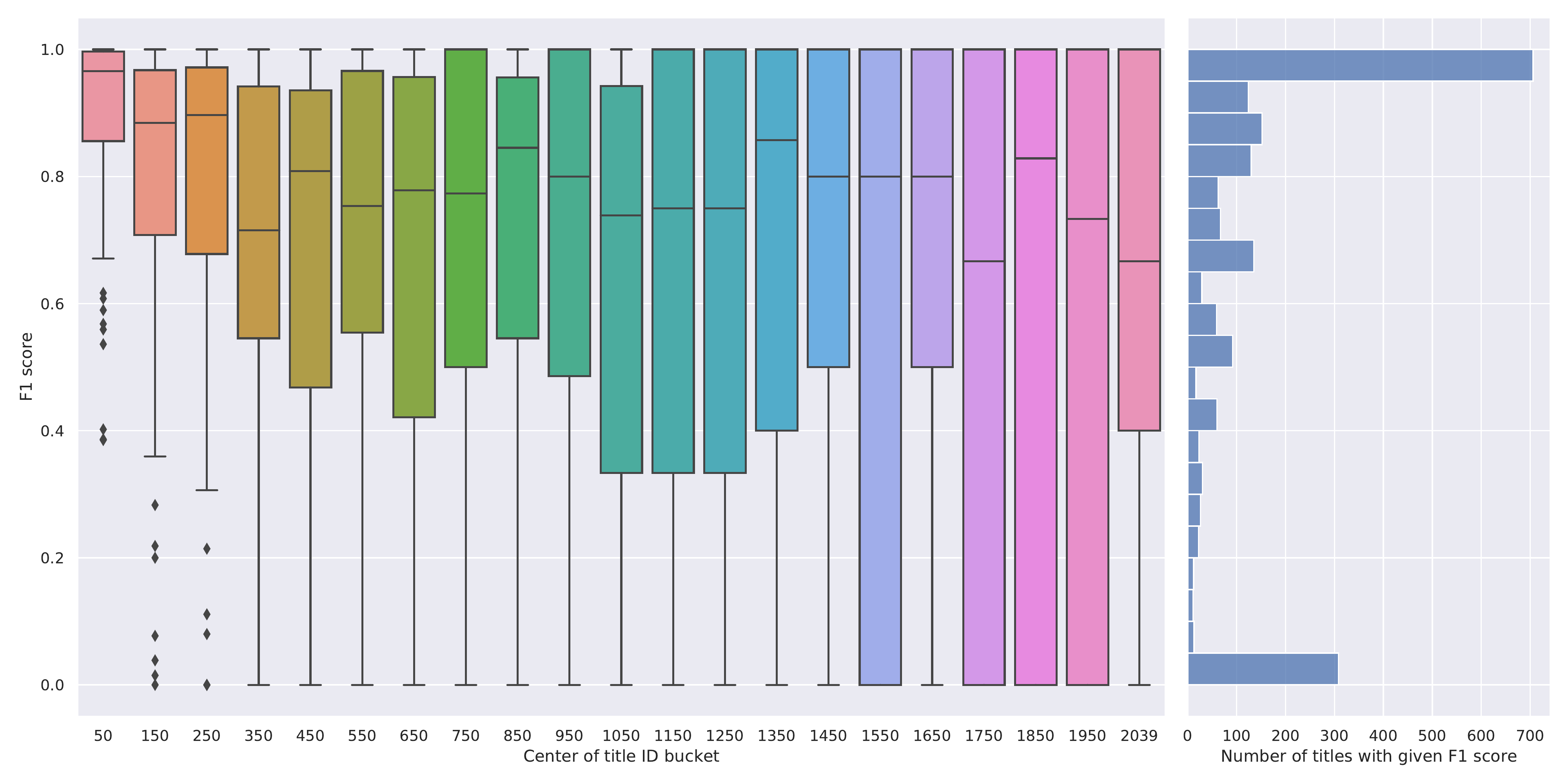}
\caption{Distribution of F1 scores of labels on the dataset with titles removed. \textit{Left:} each box represents 100 labels (apart from the last one). Title IDs used are the same as in Figure \ref{fig:label-distribution} -- they are sorted in descending order of occurrence. \textit{Right:} Histogram of F1 scores.}
\label{fig:f1-boxes}
\end{figure*}

As we can see, both Bi-LSTM and RobeCzech models achieved a very good performance. Even though RobeCzech is slightly better, it is also more than 10 times bigger and slower so in some applications it might be beneficial to use the faster Bi-LSTM architecture.

With these models we were able to achieve a surprisingly high Macro F1 score despite the long tail of the title distribution (Figure \ref{fig:label-distribution} -- some titles appear only 16 times in the train set and 4 times in the test set).

An interesting result is that LSA performs a lot better then Doc2Vec.

As was expected, the models which are able to predict based on each token separately (Bi-LSTM, RobeCzech) achieve almost 100\% accuracy on the first part of the test set (which includes segment with titles in text) (Table \ref{tab:classification-results-titles}). Even though this looks suspicious it is the intended behavior.

In the second part of the test set (segments from which we removed the titles) the metrics still look reasonable (Table \ref{tab:classification-results-notitles}) and indicate that the model is useful in further applications.

Figure \ref{fig:f1-boxes} shows detailed statistics of F1 scores on the second part (without titles) of our test set. The \textit{left plot} shows how the distribution of each bucket of 100 labels (titles) looks like. Title IDs used are the same as in Figure \ref{fig:label-distribution} -- they are sorted in descending order of occurrence, meaning the left most bucket contains F1 scores of the first 100 most frequent titles (which cover 81\% of labeled segments).
The \textit{right plot} is a simple histogram of all the F1 scores on the second part (without titles) of our test set.

We can see that around 300 titles (15\% of titles) have basically 0 F1 score. However this is not necessarily a bad thing, because their extracted title might be just an infrequent variant of a different title or a typo. We provide more details at the end of this section.

\begin{figure*}[h]
\setlength{\dbltextfloatsep}{2.0pt plus 2.0pt minus 2.0pt}
\centering
\includegraphics[width=\linewidth]{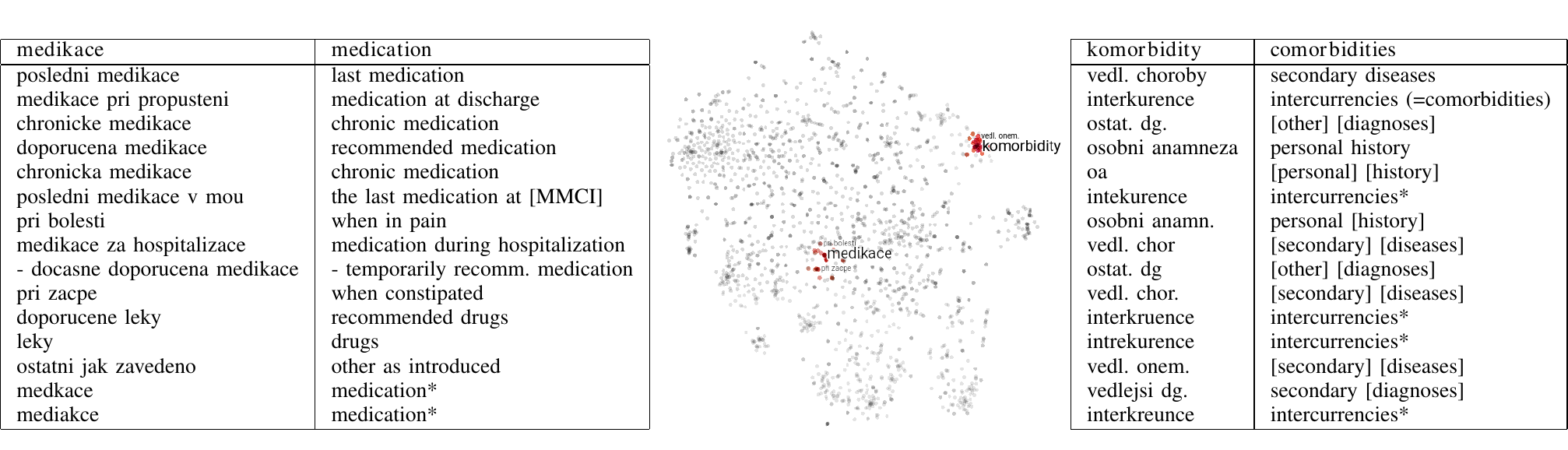}
\caption{
\textit{Center:} Nearest neighbors of the \textbf{Medication} and \textbf{Comorbidities} titles inside TensorFlow Embedding Projector with 2D t-SNE. \textit{Sides:} The 15 most similar titles to the \textbf{Medication}(left) and \textbf{Comorbidities}(right) title with manual English translation. [] -- abbreviation expansion (or translation), * -- indicates a typo in the corresponding clinical note title
}
\label{fig:title-sims}
\end{figure*}

\vspace{0.5\baselineskip}
We also performed a preliminary manual review of predictions of the RobeCzech model for segments from which we were not able to extract titles.

Our segmentation algorithm splits some types of parts of medical notes (e.g. \textit{Recommendation}, \textit{Status localis}, \textit{Objective evaluation}, \textit{Subjective evaluation}, \textit{Conclusion from genetics}, \textit{comorbidities}\dots) into multiple segments, because they are long and contain empty lines. The title is included only in the first segment thus  our models are trained only on the first segments.
Nevertheless in around 75\% of cases the RobeCzech classifier correctly predicts the title for the continuing segments. The performance is even better when we take into consideration the first three predictions.

We noticed that the model sometimes struggles with short segments (2 or 3 words). There simply might be just too little information for better predictions.

Finally we looked at frequent titles which had low F1 score on the second half of the test set (with titles removed). We~noticed two common pitfalls.

First, when the segment content often consists only from short answers (most commonly \textit{no}, \textit{negates}, \textit{sine}, \textit{0})
. This could be remedied by grouping all such segments into a new category (e.g. \textit{Unknown negation}) because it is not possible to predict the title from the content alone. Nevertheless this should not be a problem in practise because such segments are just artifacts of our pre-processing approach and should never appear in real data (even the medical professionals would not be able to make sense of it). 

Second, when there are more equivalent titles (abbreviations, typing errors) (e.g. \textit{Allergies} vs. \textit{AA}) or the same information appears in different contexts with different titles like \textit{ECG Results} in the clinicians' notes and \textit{Conclusions} in the ECG report. It is not currently possible for the classifier to distinguish between these contexts and thus it ``randomly'' chooses between them. If we had some annotations of the medical note types, we could provide them as additional features to the classifier and possibly avoid this issue.

\subsection{Clustering Pipeline}

We took the same Doc2Vec model we trained for segment classification and used the resulting document (i.e. title) embeddings with dimension 50. 

\subsubsection{Title similarity}
We computed the distance matrix from the title vector space and used it to create an ordered list of similar (cosine similarity) titles for each title.

Then we selected some pivot titles (like comorbidities or medication) and looked at their nearest neighbours.

\subsubsection{Title clustering}
We applied various clustering methods to the title embeddings, like KMeans, Agglomerative clustering, or DBSCAN. We found out that for our dataset 20 clusters yields the best results. We analysed the clusters as lists as well as in the TensorFlow Embedding Projector.\footnote{\href{https://projector.tensorflow.org/}{https://projector.tensorflow.org/}}

\subsection{Clustering Results}

\subsubsection{Similar titles}

Even though the Doc2Vec-based title classifier was not very good, it certainly has potential for computing title similarity. 

Figure \ref{fig:title-sims} shows the 15 nearest neighbours (using cosine similarity) in the Doc2Vec title vector space for the \textit{comorbidities} and \textit{medication} titles respectively. 
The Czech titles (left column) are normalized, hence they do not contain any capitalization or accents. We included manual English translation for clarity (right column). 

Figures \ref{fig:title-sims} also shows the same titles and their neighbors in the TensorFlow Embedding Projector. We used the t-SNE \cite{tsne} dimensionality reduction technique to project the title vector space into 2D. An Interactive version of this visualisation is available on GitHub.\footnote{\href{https://zepzep.github.io/clinical-notes-extraction/pages/projector/}{https://zepzep.github.io/clinical-notes-extraction/pages/projector/}} (see bookmarks at the bottom right)

As we can see these neighbours are sensible and would be useful e.g. for filtering, integrating or summarising the medical notes, or for computer assisted ontology population.

It would be difficult to come up with them based only on the title text alone, even with translation / synonym models.




\subsubsection{Title clustering}

Figure \ref{fig:title-clustering} provided at the very beginning shows an example of what sort of data we get after applying a clustering algorithm. Concretely this is the result of applying KMeans with $k=20$ to the Doc2Vec embeddings.  We labeled some clusters with their manually translated most frequent titles. Some colors are used for multiple different clusters.



\section{Discussion}



On top of the classification evaluation and preliminary cluster analysis presented in the previous section, we also performed an informal conceptual validation of the whole approach with a clinical expert from the MMCI hospital. The key points of her feedback are summarised in the following paragraphs.

First of all, the expert appreciated the results of the presented system as an important step toward the unification of clinical note segments written by different clinicians in their own personal styles. The subjective nature of the reporting process often makes it hard even for human experts to find the segment of a clinical note that contains the information they need while treating a patient.

This problem is still more pertinent to legacy clinical notes from different hospitals that often use their own information systems for structuring their records. Being able to quickly find the right segment in such ``foreign'' records is a critical clinical need in the context of oncology, since cancer patients are often transferred to specialised hospitals like MMCI after being diagnosed and initially treated elsewhere.

The accuracy acquired by our approach to segmentation and classification of Czech clinical notes was deemed by the clinical expert to be sufficient for pilot deployment in downstream tasks like patient record integration, patient similarity computation or clinical note summarisation based on the extracted topics. Last but not least, the expert also suggested the technique may be useful for more structured and standardised presentation of their own records to the patients, which would be an additional practical benefit. This is very encouraging for our future work, as elaborated in the next, final section. 

\section{Conclusion}

We have shown that it is possible and practically relevant to bring more structure to Czech unstructured clinical notes in an unsupervised manner. Our system is able to  segment clinical notes on breast cancer patients and then reliably classify the segments into categories extracted from the notes themselves.

The performance of our best classifiers is adequate for use in the downstream tasks even though there are still areas where they can be improved.

So far the analysis of our clustering techniques is anecdotal but it shows promising results. We will be able to fully evaluate the quality of generated clusters only in the context of downstream tasks we currently work on with the MMCI clinicians.

\subsection{Future work}
In future, we want to use this system as a basis for improving our baseline patient similarity model which we use for transferring personalised information between similar patient profiles. As there are many different dimensions to patient similarity (age, diagnosis, comorbidities, medication, socio-economic background, etc.) we want to filter the medical notes to include just selected segments, and only then apply the similarity calculations. We hope that by doing this we would be able to decouple individual dimensions of patient similarity and thus improve the specificity and usefulness of our predictions.

We are doing this as part of our effort to fulfill the unmet information needs of cancer patients~\cite{jenkins2001information}. Under the auspices of the AIcope project\footnote{\href{https://www.fi.muni.cz/app/projects?project=60489\&lang=en}{https://www.fi.muni.cz/app/projects?project=60489\&lang=en}}, we are actively working with clinical professionals as well as with cancer patients in order to maximise and validate the utility of our clinical information extraction, processing and presentation modules.

The filtering feature of our system can also be used to help clinicians search for relevant information faster, which is another so far unmet clinical need.

In order to effectively reduce the number of labels we want to explore different clustering methods, some of them based on the trained classifiers.

We would also like to improve the segmentation method so that it can handle multiple segments per line or multi-line segments better.

Finally we want to apply the same method to different datasets and languages to demonstrate the universality of our approach.

\bibliographystyle{IEEEtran}
\bibliography{bibliography}

\begin{thebibliography}{10}
\providecommand{\url}[1]{#1}
\csname url@samestyle\endcsname
\providecommand{\newblock}{\relax}
\providecommand{\bibinfo}[2]{#2}
\providecommand{\BIBentrySTDinterwordspacing}{\spaceskip=0pt\relax}
\providecommand{\BIBentryALTinterwordstretchfactor}{4}
\providecommand{\BIBentryALTinterwordspacing}{\spaceskip=\fontdimen2\font plus
\BIBentryALTinterwordstretchfactor\fontdimen3\font minus
  \fontdimen4\font\relax}
\providecommand{\BIBforeignlanguage}[2]{{%
\expandafter\ifx\csname l@#1\endcsname\relax
\typeout{** WARNING: IEEEtran.bst: No hyphenation pattern has been}%
\typeout{** loaded for the language `#1'. Using the pattern for}%
\typeout{** the default language instead.}%
\else
\language=\csname l@#1\endcsname
\fi
#2}}
\providecommand{\BIBdecl}{\relax}
\BIBdecl

\bibitem{kassner2021multilingual}
N.~Kassner, P.~Dufter, and H.~Sch{\"u}tze, ``Multilingual lama: Investigating
  knowledge in multilingual pretrained language models,'' \emph{arXiv preprint
  arXiv:2102.00894}, 2021.

\bibitem{li2010section}
Y.~Li, S.~Lipsky~Gorman, and N.~Elhadad, ``Section classification in clinical
  notes using supervised hidden markov model,'' in \emph{Proceedings of the 1st
  ACM International Health Informatics Symposium}, 2010, pp. 744--750.

\bibitem{ganesan2014general}
K.~Ganesan and M.~Subotin, ``A general supervised approach to segmentation of
  clinical texts,'' in \emph{2014 IEEE International Conference on Big Data
  (Big Data)}.\hskip 1em plus 0.5em minus 0.4em\relax IEEE, 2014, pp. 33--40.

\bibitem{edinger2017evaluation}
T.~Edinger, D.~Demner-Fushman, A.~M. Cohen, S.~Bedrick, and W.~Hersh,
  ``Evaluation of clinical text segmentation to facilitate cohort retrieval,''
  in \emph{AMIA Annual Symposium Proceedings}, vol. 2017.\hskip 1em plus 0.5em
  minus 0.4em\relax American Medical Informatics Association, 2017, p. 660.

\bibitem{ruan2018boundary}
W.~Ruan and W.-s. Lee, ``Boundary detection by determining the difference of
  classification probabilities of sequences: Topic segmentation of clinical
  notes,'' in \emph{2018 IEEE International Conference on Bioinformatics and
  Biomedicine (BIBM)}.\hskip 1em plus 0.5em minus 0.4em\relax IEEE, 2018, pp.
  747--750.

\bibitem{wen2021mining}
Z.~Wen, P.~Nair, C.-Y. Deng, X.~H. Lu, E.~Moseley, N.~George, C.~Lindvall, and
  Y.~Li, ``Mining heterogeneous clinical notes by multi-modal latent topic
  model,'' \emph{PloS one}, vol.~16, no.~4, p. e0249622, 2021.

\bibitem{alicante2014unsupervised}
A.~Alicante, A.~Corazza, F.~Isgr{\`o}, and S.~Silvestri, ``Unsupervised
  information extraction from italian clinical records,'' in \emph{Innovation
  in Medicine and Healthcare 2014}.\hskip 1em plus 0.5em minus 0.4em\relax IOS
  Press, 2014, pp. 340--349.

\bibitem{fette2012information}
G.~Fette, M.~Ertl, A.~W{\"o}rner, P.~Kluegl, S.~St{\"o}rk, and F.~Puppe,
  ``Information extraction from unstructured electronic health records and
  integration into a data warehouse,'' \emph{Informatik 2012}, 2012.

\bibitem{weng2017medical}
W.-H. Weng, K.~B. Wagholikar, A.~T. McCray, P.~Szolovits, and H.~C. Chueh,
  ``Medical subdomain classification of clinical notes using a machine
  learning-based natural language processing approach,'' \emph{BMC medical
  informatics and decision making}, vol.~17, no.~1, pp. 1--13, 2017.

\bibitem{patrick2011knowledge}
J.~D. Patrick, D.~H. Nguyen, Y.~Wang, and M.~Li, ``A knowledge discovery and
  reuse pipeline for information extraction in clinical notes,'' \emph{Journal
  of the american medical informatics association}, vol.~18, no.~5, pp.
  574--579, 2011.

\bibitem{lsa}
T.~K. Landauer, P.~W. Foltz, and D.~Laham, ``An introduction to latent semantic
  analysis,'' \emph{Discourse processes}, vol.~25, no. 2-3, pp. 259--284, 1998.

\bibitem{doc2vec}
\BIBentryALTinterwordspacing
Q.~V. Le and T.~Mikolov, ``Distributed representations of sentences and
  documents,'' \emph{CoRR}, vol. abs/1405.4053, 2014. [Online]. Available:
  \url{http://arxiv.org/abs/1405.4053}
\BIBentrySTDinterwordspacing

\bibitem{bilstm}
A.~Graves and J.~Schmidhuber, ``Framewise phoneme classification with
  bidirectional lstm networks,'' in \emph{Proceedings. 2005 IEEE International
  Joint Conference on Neural Networks, 2005.}, vol.~4, 2005, pp. 2047--2052
  vol. 4.

\bibitem{robeczech}
\BIBentryALTinterwordspacing
M.~Straka, J.~N{\'{a}}plava, J.~Strakov{\'{a}}, and D.~Samuel, ``{RobeCzech}:
  Czech {RoBERTa}, a monolingual contextualized language representation
  model,'' in \emph{Text, Speech, and Dialogue}.\hskip 1em plus 0.5em minus
  0.4em\relax Springer International Publishing, 2021, pp. 197--209. [Online].
  Available: \url{https://doi.org/10.1007%2F978-3-030-83527-9_17}
\BIBentrySTDinterwordspacing

\bibitem{roberta}
\BIBentryALTinterwordspacing
Y.~Liu, M.~Ott, N.~Goyal, J.~Du, M.~Joshi, D.~Chen, O.~Levy, M.~Lewis,
  L.~Zettlemoyer, and V.~Stoyanov, ``Roberta: {A} robustly optimized {BERT}
  pretraining approach,'' \emph{CoRR}, vol. abs/1907.11692, 2019. [Online].
  Available: \url{http://arxiv.org/abs/1907.11692}
\BIBentrySTDinterwordspacing

\bibitem{tsne}
\BIBentryALTinterwordspacing
L.~van~der Maaten and G.~Hinton, ``Visualizing data using t-sne,''
  \emph{Journal of machine learning research}, vol.~9, no.~86, pp. 2579--2605,
  2008. [Online]. Available:
  \url{http://jmlr.org/papers/v9/vandermaaten08a.html}
\BIBentrySTDinterwordspacing

\bibitem{jenkins2001information}
V.~Jenkins, L.~Fallowfield, and J.~Saul, ``Information needs of patients with
  cancer: results from a large study in uk cancer centres,'' \emph{British
  journal of cancer}, vol.~84, no.~1, pp. 48--51, 2001.

\end{thebibliography}


\end{document}